\documentclass[a4paper]{llncs}

\usepackage{amsfonts}
\usepackage{color}
\usepackage[colorinlistoftodos, textwidth=3cm]{todonotes} 
\usepackage{hyperref}
\usepackage{multirow}
\usepackage{geometry}
\author{Stefano Borgo\inst{1} \and Loris Bozzato\inst{2} 
\and Alessio Palmero Aprosio\inst{2} \and\\ Marco Rospocher\inst{2}
\and Luciano Serafini\inst{2}}
\institute{
  Laboratory for Applied Ontology, ISTC CNR,\\ 
  Via alla Cascata 56 C, 38123 Trento, Italy\\
  \and
  Fondazione Bruno Kessler,\\ 
  Via Sommarive 18, 38123 Trento, Italy\\
 \smallskip
 \email{stefano.borgo@cnr.it, \{bozzato,aprosio,rospocher,serafini\}@fbk.eu}
}

\def\coref{{\,\equiv_\textrm{c}\,}}

\makeatletter
\def\@seccntformat#1{\@ifundefined{#1@cntformat}%
   {\csname the#1\endcsname\quad}  
   {\csname #1@cntformat\endcsname}
}
\let\oldappendix\appendix 
\renewcommand\appendix{%
    \oldappendix
    \newcommand{\section@cntformat}{\appendixname~\thesection\quad}
}
\makeatother

\begin{document}
\title{On Coreferring Text-extracted Event Descriptions\\ with the aid of Ontological Reasoning}
\subtitle{Technical Report}

\maketitle 
\begin{abstract}
  Systems for automatic extraction of semantic information about events 
  from large textual resources are now available: these tools are capable to 
  generate RDF datasets 
  about text extracted events 
  and this knowledge can be used to reason over the recognized events.
  
  On the other hand, text based tasks for event recognition, 
  as for example event coreference 
  (i.e. recognizing whether two textual descriptions refer to the same event),
  do not take into account ontological information of the extracted events
  in their process. 
  
  In this paper, we propose a method to derive event coreference 
  on text extracted event data
  using semantic based rule reasoning.
  
  We demonstrate our method considering a limited (yet representative) set of event types:
  we introduce a formal analysis on their ontological properties and,
  on the base of this, we define a set of coreference criteria.
  We then implement these criteria as RDF-based reasoning rules
  to be applied on text extracted event data.
  We evaluate the effectiveness of our approach over a
  standard coreference benchmark dataset.
\end{abstract}

%
%
%
%

%
%


\keywords{Event coreference; Semantic information extraction; Event formalization}


\section{Introduction} 









Objects and events are basic ontological categories for knowledge and, thus, for text understanding. They have quite different properties, starting from the way they relate to time and space: objects are primarily related to space while events to time. Furthermore, in reading a book, the book (an object) is wholly present at any time during the reading (an event), while some temporal parts of the reading are not. These examples also show the role of the participation relation between objects and events: an object exists in time by participating in some event (the reader and the book participate in the reading event), and an event is spatially located where its participants are~\cite{ovchinnikova2010data}.

While research on knowledge extraction from text has traditionally concentrated on objects and their properties. Systems for automatic extraction of semantic information about events 
are more recent and have 
largely ignored
the events' ontological structure.
These tools (e.g., NewsReader~\cite{2016jws}, Pikes~\cite{2016tkde}) are clearly important and can be further exploited for other tasks. By generating RDF datasets 
focused on (text extracted) events, they make possible to use RDF knowledge to reason over the 
detected events, so that one can now automatically infer new facts. 
This opportunity allows, for example, to identify conflicting information within and across event knowledge, as well as to discover implicit relationships between events.

  
As said, text based tasks for event recognition, 
  as for example \emph{event coreference}
  (i.e. recognizing whether two textual descriptions refer to the same event),
  do not typically take into account ontological information of the extracted events
  in their process. 
  In this paper, we aim to show that an ontological analysis of event types~\cite{pianesi2000events} contributes positively in this endeavor. To achieve this, we propose a method to derive event coreference on text extracted event data using rule reasoning on the base of a well-founded ontological description of events
and their relations.
%
  Our proposed method goes as follows:
  \begin{itemize}
  \item 
    We first provide a formal analysis on events and event types  by isolating relevant ontological properties.
  \item
    On the base of this, we consider a set of criteria to discover (probable) coreferring events.
  \item
    We then implement these criteria as RDF-based reasoning rules
    to be applied on text extracted event data.
  \item
    Finally, we evaluate the effectiveness of our approach over a
    standard coreference benchmark dataset.
  \end{itemize}
%
  In particular, we will exemplify and evaluate our approach on a known 
  benchmark for the evaluation of event coreference, the 
	\emph{EventCorefBank (ECB)}\footnote{\url{http://adi.bejan.ro/data/ECB1.0.tar.gz}}~\cite{BejanH:10} corpus. 
    In presenting each step of our approach, we describe how it has been
	applied to the events and event types in the corpus and, finally, we show 
	an experimental evaluation of our implementation over the dataset.

  Our goal is to demonstrate that, by taking into account the ontological properties
	of events and their formalization as reasoning rules,
	it is possible to provide enough flexibility to 
	manage typically large, noisy and incorrect knowledge about text extracted events.  We insist that the ontology-based rules provide an additional means for knowledge extraction and should not be considered an alternative to other types of rules.


The main contributions of this paper can be summarized as follows: 
\begin{itemize}
\item 
   We first propose (in Section~\ref{sec:events}) a formal analysis of the ontological properties of
	 events and we define our event model.
\item 
   In Section~\ref{sec:tasks}
   we consider the event extraction task and we define a set of tasks
	 (in Section~\ref{sec:reas-tasks}) that can be solved via reasoning on the formal representation
	 in our model.
\item 
   Then, in Section~\ref{sec:coref} we consider the
	 particular task of event coreference 
	 and we provide its formal characterization.
\item 
   Finally, in Section~\ref{sec:exper}
	 we describe an implementation and evaluation of the approach
	 over the ECB corpus.
\end{itemize}
 
\section{Events and Event descriptions} 
\label{sec:events}

We start by stating a formal ontological definition of event, 
adapted from the approach followed in DOLCE~\cite{Borgo-M09HoO}:
\begin{quote}
An \emph{Event} is a happening, that is, an entity directly related to
time and (indirectly) related to space via its participants \cite{Borgo-M09HoO}.
\end{quote}
We slightly simplify this view by rephrasing the notion as follows: 
\begin{quote}
An \emph{Event} identifies a spatio-temporal region, some relevant entities in it, and what happens to these entities.
\end{quote}
As usual, we divide the happenings in types depending on the changes that occur to the participants during the time spanned by the event. Formally, we characterize events via four features that we take as ontologically necessary for an event to occur:
  \begin{enumerate}
	\item 
	  \emph{Event type} (i.e. the type of happening)
	\item
	  \emph{Participant(s)} with their thematic role in the event
      (i.e. their relevance in the event)
	\item
      \emph{Time of happening}
    \item
	  \emph{Global happening location}
  \end{enumerate}
For time and location features we might agree on a specified granularity.
Note that an event must have at least one participant.
We will typically distinguish \emph{agents} (persons, organisms etc.) Vs. \emph{non-agents} (objects and materials) as participants. However, the terms `active', `passive' and `tool' participant are borrowed from the linguistic terminology.
For specific types of events, one might also define the ``main'' participants, i.e. those that characterize and identify an event, and the ``accessory'' participants.

We use the term `role' in the ontological sense: the function played by the participant in the event (e.g. a person having the role of public officer in an arrest).

%
Moreover, we concentrate in particular on \emph{facts},
i.e., events that actually happened (and not, for example, hypothetical happenings or 
descriptions of events as reported in speeches).

\subsection{Formal analysis of event types} 

On the base of this, we can refine the four features by 
providing a characterization of event types by the following points:

\begin{description}
	\item[Ontological classification:]
	classification of the event type with respect to DOLCE 
	definition of events (i.e. \emph{accomplishments} or \emph{achievements})~\cite{Borgo-M09HoO}.
	\item[Active participants:]
	participants seen as active in the event.
	\item[Passive participants:]
	participants seen as passive in the event.
	\item[Tool participants:]
	participants seen as instruments in the
	event.
	\item[Subevents and sovraevents:]
	types of events that (possibly or necessarily) include or are included in the considered event.
	\item[Status before, during and after:]
	conditions on the participants that are verified before, during and after the event.
	\item[Relationship among events:]
	(possible or necessary) relations with other events (typically causality).
	\item[Symmetrical events:]
	(possible or necessary) events that take place symmetrically with the considered event.
	\item[Incompatible events/states:]
	(possible or necessary) events that cannot take place
	simultaneously with the considered event.
	\item[Number of participants:]
	conditions on the number of participants.
	\item[Spatial region:]
	conditions on the participants co-location in the considered event.
	\item[Temporal region:]
	conditions on the possible duration of the considered event.
	\item[Repeatability:]
	conditions on the possibility to repeat an event (w.r.t. the considered participants).
\end{description}

%
%
%

\subsection{Event types in ECB}

As a first step in using the ECB corpus in our experiments,
we applied our analysis on the ECB event types.
%
The \emph{EventCorefBank (ECB)} corpus~\cite{BejanH:10} is a known testset for event coreference.
It consists of 482 
news texts divided in 43 different topics. A total of 1744 event mentions are annotated, corresponding to 339 distint events.
We chose to work on this corpus because
it offers a small but relevant set of event types.

We chose 6 of the more representative event (facts) types in the collection
(namely: \emph{Arresting, Killing, Dying, Charging, Shooting, Attacking})
and we applied to them our formal analysis.
We note that these types of event are closely connected and can give rise to 
interesting coreference examples.
The complete results of the analysis on such event types are provided in Appendix~\ref{sec:appendix:types}.
For example, in Table~\ref{tab:killing-analysis}
we present the analysis of the \emph{Killing} event type.
We note that this initial analysis in our approach requires the most effort to be completed\footnote{We estimated that an expert needs around 20 min. for the analysis of an event type.}: on the other hand,
once this effort is completed, the information can be used to define sets of reasoning rules for different tasks and independent from the size and quality of the processed event data.
Note that, for some of the event types, their linguistic interpretation (viz. their definition provided in FrameNet) covers multiple types in the ontological classification (e.g., \emph{Arresting} defines both the instantaneous act of seizing someone as well as the whole process of searching, seizing and taking the person into custody). In the current experiment we chose only one of the interpretations (e.g. we consider \emph{Arresting} as an achievement) in order to define an univocal set of coreference rules: however, in a more complete characterization, one may define different analysis for both interpretations and infer from the text context which one is intended.

\begin{table*}[htb!]%
\caption{Formal analysis of event type \emph{Killing}}
\centering\small
\begin{tabular}{|l|p{.6\textwidth}|}
	\hline
	\textbf{Ontological classification} & Accomplishment\\
	\hline
	\textbf{Active participant} & Physical object (including agents)\\
	\hline
  \textbf{Passive participant} & Living entity\\
  \hline
  \textbf{Tool participant} & Physical object (including chemical and biological entities)\\
  \hline
  \textbf{Subevents}	& Hitting (possible); start of the killing, dying (necessary)\\
  \hline
  \textbf{Sovraevents}	& Murdering, colliding, starving, being infected (all possible)\\
  \hline
  \textbf{Status: before} & Passive participant is alive\\
  \hline
  \textbf{Status: during} & Passive participant is alive\\
  \hline
  \textbf{Status: after} & Passive participant is dead\\
  \hline
  \textbf{Relationship among events} &	Killing causes: damaging\\
  \hline
  \textbf{Symmetrical events} & Dying for causes external to the passive participant\\
  \hline
  \textbf{Incompatible events/state} & --\\
  \hline
  \textbf{Number of participants} & Killing can have one or more active participants; one or more passive participants; 
  one or more tool participants.\\
  \hline
  \textbf{Spatial region} & Killing happens in the location where the passive participants are located\\
  \hline
  \textbf{Temporal region} & Interval\\
  \hline
  \textbf{Repeatability}   & There cannot be two killing events with the same passive participant;
	there can be more than one killing event with the active participant;
	there can be more than one killing event in the same spatio-temporal region.\\	
	\hline
\end{tabular}
\label{tab:killing-analysis}
\end{table*}



\section{Event extraction from text}
\label{sec:tasks}

Information about events can be gathered from many resources. One of
the most commonly used resource for event description is
text. News article describes events, hypothetical events, contains
event prediction, etc. 

The growing maturity level of NLP techniques makes now possible to
automatic process large amount of textual resources and extract
information about the events mentioned in the text. The output however
cannot be considered as certain information, as errors might occurs in
the extraction process, or wrong information could be present in the
source text.

Typically, various types of information about events can be automatically 
extracted from a text corpus. We review them looking at the output produced by one of the
state-of-the-art tools for knowledge extraction, Pikes~\cite{2016tkde}.

Pikes is a Knowledge Extraction framework adopting a 2-phase approach. First---phase~1: \emph{linguistic feature extraction}---an RDF graph of mentions is built by distilling the output of several state-of-the-art NLP tools, including Stanford CoreNLP\footnote{\url{http://nlp.stanford.edu/software/corenlp.shtml}} (tokenization, POS-tagging, lemmatization, NERC, TERN, parsing and coreference resolution), UKB\footnote{\url{http://ixa2.si.ehu.es/ukb/}} (WSD), DBpedia Spotlight\footnote{\url{http://spotlight.dbpedia.org/}} (EL), Mate-tools\footnote{\url{http://code.google.com/p/mate-tools/}} and Semafor\footnote{\url{http://www.cs.cmu.edu/~ark/SEMAFOR/}} (SRL). 
Then---phase~2: \emph{knowledge distillation}---the mention graph is processed to distill the knowledge graph using SPARQL-like mapping rules, which are evaluated using RDFpro\footnote{\url{http://rdfpro.fbk.eu/}}~\cite{2015sac}, an RDF manipulation tool used also for RDFS reasoning.

\begin{figure}[tb]
	\centering	
	\includegraphics[width=0.80\textwidth]{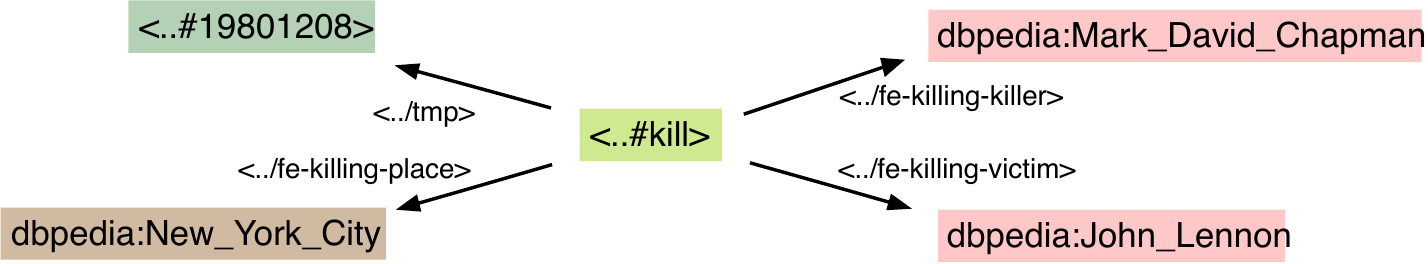}
	\caption{Graphical representation of the output produced by Pikes on the example sentence.}
	\label{fig:pikes}
\end{figure}
The RDF knowledge graph returned by Pikes contains various typologies of content, that we illustrate using as example the output produced when parsing the sentence ``\emph{On December 8, 1980, in NYC, Mark Chapman killed John Lennon.}'' (see Figure~\ref{fig:pikes} for a graphical rendering of the output produced or directly access the online demo on Pikes web-site\footnote{\url{http://pikes.fbk.eu}}):
\begin{itemize}
	\item event instances (e.g., ``kill''), typed with respect to known ontological (e.g., FrameBase\footnote{\url{http://www.framebase.org/}}, SUMO\footnote{\url{http://www.adampease.org/OP/}}) and linguistic resources (e.g., FrameNet\footnote{\url{http://framenet.icsi.berkeley.edu/}}, VerbNet\footnote{\url{https://verbs.colorado.edu/~mpalmer/projects/verbnet.html}});
	\item event participants (e.g., ``dbpedia:John\_Lennon''), i.e., individuals possibly linked to typed instances in linked data repositories (e.g., DBpedia\footnote{\url{http://dbpedia.org/}});
	\item role of participants in the event (e.g., ``killing-victim''), according to role catalogs taken from available ontological (e.g., FrameBase) and linguistic resources (e.g., FrameNet, VerbNet);
	\item temporal information (e.g., ``19801208''), grounding the event in time, formalized according to reference time ontology (e.g., OWL-time\footnote{\url{https://www.w3.org/2006/time}});
	\item location information (e.g., ``dbpedia:New\_York\_City''), grounding the event geographically.
\end{itemize}
This output perfectly fits the 4-feature ontologically necessary characterization of events introduced at the beginning of Section~\ref{sec:events}.


\section{Ontological-based reasoning\\[.3ex] tasks for events}
\label{sec:reas-tasks}
Provided the formal ontological characterization of event types presented in the 
analysis in Section~\ref{sec:events}, our goal is to demonstrate that
this knowledge enables us to deal with different high-level reasoning tasks on events 
and their properties.
In the following, we briefly exemplify some of these possible tasks and,
following our approach, we mention what kind of information from the ontological 
analysis can be used; in the remainder of the paper, we will then turn our attention to 
the specific task of event coreference.


\smallskip\noindent
\textbf{Pre- and Post-condition inference.}
One task regards the ability to reason on the states that enable
the activation of an event (i.e. its pre-conditions), the effects on the 
knowledge that the event produces on the state (i.e. its post-conditions), and possibly the 
facts that are true during the event execution (aka ``implied situations'' in~\cite{2015maplex}).
%
For example, supposing that we extract the event \emph{``John has been hired by Google''},
then we would like to be able to infer that before this event 
\emph{``John was not working in Google''} while later this fact is true.

The ontological properties that can be used for this task are clearly
the information about status before, during and after events and possibly
information on super- and sub-events with their associated knowledge content.
%

\smallskip\noindent
\textbf{Completion of missing events.}
Another reasoning task can regard the inference of missing or implicit events 
and the completion of the events set on the base of the available event information.
%
For example, suppose that we detect the two events \emph{``Jobs is the CEO of
  Apple''}, and \emph{``Cook is the CEO of Apple''} at different dates. 
We would like to infer that an event of the appointing of the new
  CEO of Apple has happened between the two events.

In this case, the useful ontological features can regard the status before, during and after the
event, the possible asserted relations across event types (e.g. causality)
and the local information about the participants.

\smallskip\noindent
\textbf{Event information refinement.}
On the base of the ontological information, we might also want to 
refine the extracted information associated to an event in order to reach a 
more fine grained representation of the event local knowledge. 
This can be achieved possibly
using contextual information known from the previously recognized events
or background knowledge.
%
For instance, suppose that we extract the event \emph{``The president visited the capital of France
and met the prime minister''} in the time period of July 2016 and American politics news,
then we might infer that \emph{``Barack Obama visited Paris and met Manuel Valls''}.

This task requires ontological information about roles and the ability to reason
with contextual information (e.g. on the time and location of the event occurence).


\smallskip\noindent
\textbf{Incompatible events.}
From the text extraction, we can obtain two events that have incompatible information
(e.g. possibly coming from different sources):
we want to be able to find that the happening is the same but the information is incompatible 
(due to errors or simply different versions of the facts).
%
For example, we might have that a news reports \emph{``Last night, John Smith killed two people with a knife''}
while another news might report \emph{``Last night, an unidentified suspect killed a woman with a knife''}:
without knowing which version of the facts actually happened, 
we want to be able to assert that the two descriptions are incompatible.

To be able to obtain these inferences, we might want to compare the event descriptions that refer to the same 
event with respect to their roles, time and location in order to recognize the incompatible
information and use information about incompatible states.

 
\section{Event coreference task} 
\label{sec:coref}

In the following, we will first provide a definition of event coreference,
then we show how the task of coreference discovery has been formalized and implemented 
in our experiments on the base of the ontological representation of events.

\subsection{Coreference definition}
In this paper, we define \emph{event coreference} as an equivalence relation 
$\coref$ among events: 
$e_1 \coref e_2$ holds when events $e_1$ and $e_2$ 
describe the same happening or parts of the same happening 
regarding the four characterizing features
described in Section~\ref{sec:events}.
In other terms, the features of $e_1$ and $e_2$ identify different classifications (perspectives) of the same happening or the set of features of $e_1$ ($e_2$) identifies a part of the happening identified by $e_2$ ($e_2$, resp.ly).

Note that even when event coreference holds between two events $e_1$ and $e_2$, these events can still differ in terms of information content. For instance, if the event is an olympic race, $e_1$ may have information about the weather during the race which might not be available in event $e_2$. Indeed, for event coreference to be satisfied it suffices that they both have the same information regarding the event type (a race), time, participants and location.



The task that we consider with respect to event coreference is 
to determine the existence and degree of truth
to which a coreference relation exists based on the related data on the events.
Our underlying hypothesis is that an analysis at the event level, 
based on the proposed ontological properties and linguistic resources (e.g. FrameNet~\cite{Baker:98}), 
can emend and complete the information extracted from single event detection.

\subsection{Coreference measure and aggregation} 

We can formally define a coreference measure and a relation to aggregate event descriptions as follows.

We call a \emph{coreference measure} $\mu_{t}$ for event type $t$ any function that 
assigns to every pair of event descriptions $ed_1,ed_2$ of type $t$ 
a value in $[0,1]$.
%
Intuitively a coreference measure, provides an estimation of the plausibility for two event descriptions to refer to the same event.

A \emph{coreference aggregation} $\alpha_{t}$ for event type $t$
is a function that given two event descriptions, $ed_1,ed_2$ of type $t$
returns a new joint event description $ed_{1+2}$. 
A coreference aggregation provides a way to merge two event descriptions into a single event 
description that aggregates all the information contained in the starting event descriptions.


\subsection{Implementing coreference conditions rules}

In our approach, rules for coreference measure and aggregation
are formulated starting from the ontological 
definition of events.

%
In our first formulation for coreference rules 
we only considered the conditions for coreference measure:
that is, in our experiments we are not interested in aggregating 
event descriptions but we only recognize and mark them as (possibly) coreferring.

As a first formulation for coreference measure, 
we simply provided rules for \emph{certain} and \emph{possible} coreference:
the first conditions correspond to sufficient and necessary
conditions (for each event type) for coreference existence,
while the latter constitute only necessary conditions,
providing an evidence for possible existence of coreference.
Such conditions have been extracted from the ontological
analysis of the considered event type: in other words,
these coreference criteria correspond to necessary (and sometime sufficient) 
event identity criteria that result from that analysis.

In the formulation of the rules,
we considered the availability of FrameNet frame elements data and the 
event type properties defined in the ontological analysis.
These rules have then been implemented as SPARQL rules and used in the evaluation
of our approach, as described in next section.

The complete set of rules can be found in Appendix~\ref{sec:appendix:rules}.
In Table~\ref{tab:killing-conditions}, for example, we report the 
conditions we identified for the event type \emph{Killing}.
In the table, each row represents one condition: $E_1$ and $E_2$ are variables identifying two \emph{Killing} event instances and we use the dot notation to denote the value of their properties (viz. frame elements). With $==$ we indicate equality of the values and with $\sim$ their compatibility (e.g. time value of $E_1$ can be included of the time value for $E_2$). The relation $hasCoref$ indicates that the compared (sub-)events are themselves recognized as coreferring.

\begin{table*}[ht]%
\caption{Conditions for event type \emph{Killing}}
\centering
  \begin{tabular}{|l|c|}
  \hline
	\multirow{2}{*}{certain coref.} 
                  & $(E1.Victim == E2.Victim)$ \\[1ex]
	              & $(E1.SubEvent\ hasCoref\ E2.SubEvent)$\\
  \hline\mbox{}\\[-1.5ex]
  \multirow{5}{*}{possible coref.} 
                  & $(E1.Killer == E2.Killer)\ \&\ (E1.Time \sim E2.Time)$\\[1ex]
                  & $(E1.Killer == E2.Killer)\ \&\ (E1.Place \sim E2.Place)$\\[1ex]									
                  & $(E1.Killer == E2.Killer)\ \&\ (E1.Tool == E2.Tool)$\\[1ex]																		
                  & $(E1.Tool == E2.Tool)\ \&\ (E1.Time \sim E2.Time)$\\[1ex]
                  & $(E1.Place \sim E2.Place)\ \&\ (E1.Time \sim E2.Time)$\\
  \hline
  \end{tabular}
\label{tab:killing-conditions}
\end{table*}
\noindent
We remark that in this version of the rules we only considered ``intra-type''
coreference rules: this choice was made in order to compare with the existing 
coreference approaches in the evaluation.
However, it is easy to extend the set of rules with conditions establishing 
coreference relations across events of different types. 
For example, considering \emph{Killing} and \emph{Dying} events,
we can assert that they (certainly) describe the same happening if the victim is the same
individual: we can write this condition as 
\begin{center}\small
	$(E1.Victim == E2.Protagonist)\ \&\ Killing(E1)\ \&\ Dying(E2)$
\end{center}



\section{Evaluation on ECB benchmark} 
\label{sec:exper}

In this section, we evaluate the extraction of the coreference on the extended Event Coreference Bank corpus \cite{BejanH:10}.
The dataset is a collection of 482 text documents from Google News, divided into 43 sets representing different topics.
For each topic, the set of texts is annotated with intra- and cross-document event coreference, in accordance with the TimeML specification.

In literature, there are different ways to evaluate event coreference, each of them having different peculiarity: e.g., MUC~\cite{Vilain:1995:MCS:1072399.1072405}, B$^3$~\cite{Bagga98algorithmsfor}, CEAF~\cite{Luo:2005:CRP:1220575.1220579}, and BLANC~\cite{42559}.
BLANC, in particular, was developed to overcome the limitations of the other previously proposed metrics, as it takes into consideration both coreference and non-coreference links, does not ignore singletons (problematic for MUC) and 
does not boost the score in their presence (as typically occurs for B$^3$ and CEAF scores).

To compute these metrics, we relied on an already available package, CorScorer\footnote{
\url{http://conll.github.io/reference-coreference-scorers/}}, an open source tool that measure coreference sets w.r.t. the most important metrics used for this kind of evaluation.

\subsection{Experiments setup}

Unfortunately, not every event in the ECB corpus is annotated in the document, therefore we choose to consider, in our classification, only the words that are actually annotated in the gold standard.

In addition, as we are currently dealing with the most-frequent 6 event types, we filter out the lemmas that do not appear as lexical units in the corresponding frames in FrameNet.
In particular, our evaluation is restricted to 137 different lemmas.

We develop a lemma-based baseline to be compared with the results obtained by our system: all predicates in FrameNet that have the same lemma are coreferential.

We processed the ECB corpus with Pikes, obtaining an RDF knowledge graph of the knowledge about events, entities, locations, etc., conveyed by the news documents in the corpus. We then applied the rules expressing coreference conditions on this RDF knowledge graph. The rules were implemented as SPARQL based rules, enriching the input knowledge graph with additional \texttt{possibleCoref} or \texttt{certainCoref} triples based on the condition expressed in the body. The rules were applied to the RDF data using RDFpro~\cite{2015sac}. 

\subsection{Experiments results}

Results of our experiments are shown in Table~\ref{tab:results}. Besides the baseline, we report the performances of applying possible and certain event coreference rules separately, and altogether. A total of 152 possible event coreference pairs and 275 certain event coreference pairs were generated.
\begin{table*}
\centering
	\scriptsize
	\caption{Experiments Results (numbers are percentages)}	
	\label{tab:results}    
	\begin{tabular}{l|r|r|r|r|r|r|r|r|r|r|r|r|r|r|r}
        & \multicolumn{3}{c|}{MUC} & \multicolumn{3}{c|}{B$^3$}  & \multicolumn{3}{c|}{CEAF (M)} & \multicolumn{3}{c}{BLANC} \\ \hline \hline
& p    & r & F1 & p        & r & F1 & p      & r & F1 & p       & r & F1 \\ \hline \hline
		lemma baseline     & 81.78    & 89.14 & 85.30  & 39.87        & 73.46 & 51.69  & 45.48      & 45.48 & 45.48  & 62.45     & 76.08 & 66.14  \\ \hline
		only certain       & 100    & 7.39 & 13.77  & 100        & 2.76 & 5.38  & 86.34      & 6.96 & 12.88   & 96.47    & 88.38 & 91.6  \\ \hline
		only possible      & 100    & 6.08 & 11.46  & 100        & 3.35 & 6.5  & 82.14      & 5.84 & 10.91   & 95.07     & 90.54 & 92.18  \\ \hline
		possible + certain & 100    & 13.48 & 23.75  & 100        & 6.12 & 11.54  & 84.37      & 12.8 & 22.23  &  98.23     & 89.6 & 93.29  \\ \hline
	\end{tabular}
\end{table*}
On BLANC, our approach clearly outperforms the lemma baseline (approx. 27\% difference on F1, considering both probable and certain coreference). On all metrics, our work consistently shows better precision scores than the baseline, while the contrary holds for recall. The good performances on precision should not surprise, as our rules get fired only when several conditions on the event description are met, and works also on events of the same type but expressed with different lemmas. On the contrary, the low performances on recall, especially compared with the baselines, are partly justified by the nature of the dataset considered in the evaluation: it consists of 43 different topics, and within each topic all documents mostly refer to the very same event, thus an approach based on matching lemma predicates it is likely to show good overall performances. However, in a general event extraction context, especially when dealing with event extraction from large news documents, such baseline approach will likely corefer very different event mentions just because they share the same predicate. In such context, precision-tuned approaches, as our work, tend to be more successful as they better cope with the noise and redundancy of information. As a final remark, our approach performs better when considering both certain and possible rules.
\section{Related work} 

Determining when two event descriptions in a text corpus are about the same event is a challenging, popular research stream.
Several approaches have been proposed, and we summarize the main ones in this section.
%
Approaches typically exploit the event arguments (e.g., action, participants, location, time) when comparing and clustering events descriptions (cfr.~\cite{2016kbs}).

Among them, some approaches \cite{humphreys-etal-1997,ChenJi-2009a} adopt models consisting of labeled training data. \cite{humphreys-etal-1997} proposed an ontology-based deterministic clustering algorithm to group event descriptions of certain type. \cite{ChenJi-2009a} presented a learning-based classification algorithm exploiting event arguments and features.

In~\cite{BejanH:10} an unsupervised Bayesian clustering model is proposed for within- and cross-document event coreference. It is based on the hierarchical Dirichlet process, and 
grounded on the assumption that two event descriptions corefer if they have the same event properties and share the same event participants. 

\cite{Lee+etal+2012} presented an iterative approach, where entity coreference and event coreference are jointly performed, so that at each iteration, the intermediate results produced for one of the two tasks can be used in solving the other.

In~\cite{Liu-etal-2014} a supervised method for event coreference is proposed: the method exploits a rich feature set, and propagates information (alternatively) between events and their arguments.

Finally, \cite{cybulska2015bag} considered granularity (e.g., durations of event actions, granularity level of locations) in computing event coreference. The authors present a supervised pairwise binary classifier based on decision-trees which computes the compatibility of event attributes (e.g., event trigger, time, location, human and non-human participants).

Notwithstanding some attempts to use ontologies (e.g.~\cite{humphreys-etal-1997}), none of the state-of-the-art approaches for within- and cross-document event coreference makes use of an ontological representation of events as presented in this paper.

On the other side,
there is a long history of proposals for ontological definitions of events, for example inside DOLCE~\cite{Borgo-M09HoO},
conceptualization efforts as the \emph{Event Model F}~\cite{ScherpFSS:09} or ontology design patterns\footnote{See, 
for example, \url{http://ontologydesignpatterns.org/wiki/Submissions:EventCore}}:
in the context of Semantic Web data, some works considered 
reasoning with ontologically well founded definitions of events in a noisy and 
incomplete knowledge scenario like the case of (text extracted) news data.
%
For instance,~\cite{MeleS:11} introduces a formalization of composite events 
through a formal ontology called \emph{Ontology of Complex Events (OntoCE)}.
This work aims at integrating complex event information from different media
by managing incompatibilities across their contents. Such definition of events is then used to reason 
for checking temporal consistency of complex events,
find new temporal relations and perform causal reasoning.
On the other hand, in events extraction, ontological models (like the ESO ontology~\cite{2015maplex}) 
are generally used to structure and expose events objects and reasoning is only applied at a high level,
assuming already clean and reconciled representations.

Other works do not aim at a foundational characterization of event types,
but concentrate on defining the properties of events
in particular domains and (textual) resources. 
For example, \cite{byrne2010automatic} focus on characterizing text extracted events  
for RDF representation and reasoning over archeological procedures.
Similarly, \cite{Rovera:16} 
outlines a project of an ontology-based framework to formally
represent and extract historical domain events from archives in form of linked data: 
purpose of this extraction is to semi-automatically build narratives of complex events.

In the area of event representation, we may also cite works regarding
Semantic Web enabled event stream reasoning: for example, 
\cite{keskisarkka2013semantic} surveys the features of different approaches
for analyzing streams of social media data to extract complex events.
The authors, however, notice that the current approaches are limited by
non-realistic assumptions of well-structured data, known streams and 
static ontologies. Current approaches lack 
support for temporal and spatial reasoning on complex events and 
usually do not support RDFS or OWL reasoning, 
restricting the possibilities to use background knowledge (thus including
ontological definitions of events).


As noticed, the proposed approaches are usually interested in reasoning on high
level (ontological) features and classification of events: in our work, our goal
is to demonstrate that also ``low level'' processing tasks as event coreference
can benefit from a well-founded characterization of events.
Moreover, the contribution of our work does not reside in showing the benefits
of using an event ontology for the execution of such tasks, but in the application
of a prior ontological analysis of events types present in the resources.

\section{Conclusion}
In this paper we presented a method to derive coreference of
text-extracted events on the base of a well-founded ontological analysis 
of events and their relations.
We first provided a formal definition of event and presented a form of 
ontological analysis of event types.
On the base of this, we defined a set of conditions for 
the discovery of event coreference, which we easily implemented as RDF-based reasoning rules.
Finally, we evaluated our approach and hypotheses over the ECB corpus:
the experiments show promising results encouraging us for further investigation in the direction of this initial work.

In this regard, we plan on extending the current work by considering a larger set of event types to be analyzed and the possiblity to apply our method to larger news datasets. Moreover, we also want to apply the method to different event reasoning tasks, like the ones identified in Section~\ref{sec:reas-tasks}.

\appendix
\newpage
\section{Analysis of Event Types}
\label{sec:appendix:types}
\subsection{Arresting}

\begin{tabular}{|l|p{.6\textwidth}|}
	\hline
	\textbf{Ontological classification} & Achievement\\
	\hline
	\textbf{Active participant} & Person (role: public officer)\\
	\hline
  \textbf{Passive participant} & Person (role: civil agent)\\
  \hline
  \textbf{Tool participant} & Law\\
  \hline
  \textbf{Subevents}	& --\\
  \hline
  \textbf{Sovraevents}	& Declaration of arrest\\
  \hline
  \textbf{Status: before} & Passive participant is alive\\
  \hline
  \textbf{Status: during} & Non applicable\\
  \hline
  \textbf{Status: after} & Passive participant is alive and subject to (legal) restrictions\\
  \hline
  \textbf{Relationship among events} & Arrest causes: being into legal custody  (necessary); being prisoner  (necessary); being handcuffed (possible)\\
  \hline
  \textbf{Symmetrical events} & To give up rights to the legal authority\\
  \hline
  \textbf{Incompatible events/state} & Events in which an active or passive participant of this event is released, dead or unconscious; a simultaneous arrest event in which the passive participant of this event is an active participant of the other (and vice versa)\\
  \hline
  \textbf{Number of participants} & Arrest must have: one or more active participants; one or more passive participants; one or more tool participants\\
\hline
  \textbf{Spatial region} & Arrest happens in the location where the active and passive participants are located\\
  \hline
  \textbf{Temporal region} & Atomic\\
  \hline
  \textbf{Repeatability} & There can be several arrest events with the same passive participants;
there can be more than one arrest event with the same active participants;
there can be more than one arrest event in the same spatio-temporal region\\	
	\hline
\end{tabular}

\subsection{Killing}

\begin{tabular}{|l|p{.6\textwidth}|}
	\hline
	\textbf{Ontological classification} & Accomplishment\\
	\hline
	\textbf{Active participant} & Physical object (including agents)\\
	\hline
  \textbf{Passive participant} & Living entity\\
  \hline
  \textbf{Tool participant} & Physical object (including chemical and biological entities)\\
  \hline
  \textbf{Subevents}	& Hitting (possible); start of the killing, dying (necessary) \\
  \hline
  \textbf{Sovraevents}	& Murdering, starving, being infected (all possible)\\
  \hline
  \textbf{Status: before} & Passive participant is alive\\
  \hline
  \textbf{Status: during} & Passive participant is alive\\
  \hline
  \textbf{Status: after} & Passive participant is dead\\
  \hline
  \textbf{Relationship among events} &	Killing causes: damaging\\
  \hline
  \textbf{Symmetrical events} & Dying for causes external to the passive participant\\
  \hline
  \textbf{Incompatible events/state} & --\\
  \hline
  \textbf{Number of participants} & Killing can have one or more active participants; one or more passive participants; 
  one or more tool participants.\\
  \hline
  \textbf{Spatial region} & Killing happens in the location where the passive participants are located\\
  \hline
  \textbf{Temporal region} & Interval\\
  \hline
  \textbf{Repeatability}	& There cannot be two killing events with the same passive participant;
	there can be more than one killing event with the same active participant;
	there can be more than one killing event in the same spatio-temporal region.\\	
	\hline
\end{tabular}

\subsection{Dying}

\begin{tabular}{|l|p{.6\textwidth}|}
	\hline
	\textbf{Ontological classification} & Achievement\\
	\hline
	\textbf{Active participant} & --\\
	\hline
  \textbf{Passive participant} & Living entity\\
  \hline
  \textbf{Tool participant} & Physical object (including chemical and biological entities)\\
  \hline
  \textbf{Subevents}	& --\\
  \hline
  \textbf{Sovraevents}	& Killing\\
  \hline
  \textbf{Status: before} & Passive participant is alive\\
  \hline
  \textbf{Status: during} & Non applicable\\
  \hline
  \textbf{Status: after} & Passive participant is dead\\
  \hline
  \textbf{Relationship among events} &	Dying is caused by: murdering (possible); being infected (possible); begin poisoned (possible)\\
  \hline
  \textbf{Symmetrical events} & --\\
  \hline
  \textbf{Incompatible events/state} & Being born; acting\\
  \hline
  \textbf{Number of participants} & Dying can have one or more passive participants; one or more tool participants\\
  \hline
  \textbf{Spatial region} & Dying happens in the region where the passive participant is located\\
  \hline
  \textbf{Temporal region} & Atomic\\
  \hline
  \textbf{Repeatability} & 
  There can be only one dying event with the same passive participant;
there can be only one dying event in the same spatiotemporal region.\\	
	\hline
\end{tabular}

\subsection{Charging (via legal notification)}

\begin{tabular}{|l|p{.6\textwidth}|}
	\hline
	\textbf{Ontological classification} & Achievement\\
	\hline
	\textbf{Active participant} & Person (role: public officer)\\
	\hline
  \textbf{Passive participant} & Person (role: civil agent)\\
  \hline
  \textbf{Tool participant} & Law, written text (depending on the law)\\
  \hline
  \textbf{Subevents}	& --\\
  \hline
  \textbf{Sovraevents}	& --\\
  \hline
  \textbf{Status: before} & Passive participant is alive and conscious, a judge rules the charge against the passive participant\\
  \hline
  \textbf{Status: during} & Non applicable\\
                       
  \hline
  \textbf{Status: after} & The passive participant is charged\\
  \hline
  \textbf{Relationship among events} & Charging causes: acquisition of a legal status (necessary); begin under spacial legal restrictions (possible); being arrested (possible)\\
  \hline
  \textbf{Symmetrical events} & --\\
  \hline
  \textbf{Incompatible events/state} & Being dead; being unconscious\\
  \hline
  \textbf{Number of participants} & Charging has one or more active participants; one or more passive participants, one or more tool participants\\
  \hline
  \textbf{Spatial region} & Charging happens in the region where the passive participant is located\\
  \hline
  \textbf{Temporal region} & Atomic\\
  \hline
  \textbf{Repeatability} & 
There can be several charging events with the same active and passive participants;
there can be several charging events in the same spatiotemporal region.\\	
	\hline
\end{tabular}

\subsection{Shooting}

\begin{tabular}{|l|p{.6\textwidth}|}
	\hline
	\textbf{Ontological classification} & Accomplishment\\
	\hline
	\textbf{Active participant} & Agent, physical object (gun, bow)\\
	\hline
  \textbf{Passive participant} & Physical object (bullet, arrow, stone), amount of matter (sand, water)\\
  \hline
  \textbf{Tool participant} & Physical object (depending on active and passive participants: gun for person and bullet; bow for person and arrow, non applicable for person and spear)\\
  \hline
  \textbf{Subevents}	& Exercising force, throwing, moving\\
  \hline
  \textbf{Sovraevents}	& Murdering, attacking\\
  \hline
  \textbf{Status: before} & Active participant controls the passive participant or the tool participant, the tool participant is loaded\\
  \hline
  \textbf{Status: during} & Active participant controls the passive participant or the tool participant, the tool participant is unloaded\\
  \hline
  \textbf{Status: after} & Passive participant is moving, tool participant is unloaded (possible)\\
  \hline
  \textbf{Relationship among events} & Shooting causes: moving of the passive participant (necessary)\\
  \hline
  \textbf{Symmetrical events} & --\\
  \hline
  \textbf{Incompatible events/state} & Staying still, having no propelling power, being unloaded\\
  \hline
  \textbf{Number of participants} & Shooting can have one or more active participants; one or more passive participants; one or more tool participants\\
  \hline
  \textbf{Spatial region} & Shooting happens in the region where the active and passive participants are located\\
  \hline
  \textbf{Temporal region} & Interval\\
  \hline
  \textbf{Repeatability} & 
There can be several shooting events with the same passive participant;
there can be several shooting events with the same active participant;
there can be several shooting events in the same spatiotemporal region\\	
	\hline
\end{tabular}

\subsection{Attacking}

\begin{tabular}{|l|p{.6\textwidth}|}
	\hline
	\textbf{Ontological classification} & Accomplishment\\
	\hline
	\textbf{Active participant} & Agent, biological entity (including system)\\
	\hline
  \textbf{Passive participant} & Agent, physical object, biological entity (including system)\\
  \hline
  \textbf{Tool participant} & Biological entity, physical object, amount of matter\\
  \hline
  \textbf{Subevents}	& Start of the attack; end of the attack; hitting; destroying, moving\\
  \hline
  \textbf{Sovraevents}	& Fighting, competing\\
  \hline
  \textbf{Status: before} & Active participant is alive or functioning, has control of the tool (if present)\\
  \hline
  \textbf{Status: during} & Active participant acts and controls the tool (if present)\\
  \hline
  \textbf{Status: after} & --\\
  \hline
  \textbf{Relationship among events} & Attacking causes: harming passive participant (possible); defending (possible)\\
  \hline
  \textbf{Symmetrical events} & --\\
  \hline
  \textbf{Incompatible events/state} & Being dead, being unconscious (for active participant)\\
  \hline
  \textbf{Number of participants} & Attacking can have one or more active participants; one or more passive participants; one or more tool participants\\
  \hline
  \textbf{Spatial region} & Attacking happens in the region where the tool and passive participants are located\\
  \hline
  \textbf{Temporal region} & Interval\\
  \hline
  \textbf{Repeatability} & 
there can be several attacking events with the same passive participant;
there can be several attacking events with the same active participant;
there can be several attacking events in the same spatiotemporal region\\	
	\hline
\end{tabular}
\newpage
\section{Coreference conditions}
\label{sec:appendix:rules}
\subsection{Arresting}

  \begin{tabular}{|l|c|}
  \hline
	\multirow{1}{*}{certain coref.} 
	                & $(E1.Suspect == E2.Suspect)\ \&\ (E1.Time \sim E2.Time) $\\[.5ex]
  \hline\mbox{}&\\[-2ex]
  \multirow{3}{*}{possible coref.} 
                  & $(E1.Suspect == E2.Suspect)\ \&\ (E1.Offense == E2.Offense) $\\[1ex]
                  & $(E1.Suspect == E2.Suspect)\ \&\ (E1.Place \sim E2.Place) $\\[1ex]									
                  & $(E1.Place \sim E2.Place)\ \&\ (E1.Time \sim E2.Time)\ \&$\\ & $(E1.Offense == E2.Offense)$\\[1ex]																	
  \hline
  \end{tabular}

\subsection{Killing}

  \begin{tabular}{|l|c|}
  \hline
	\multirow{2}{*}{certain coref.} 
                  & $(E1.Victim == E2.Victim)$ \\[1ex]
	              & $(E1.SubEvent\ hasCoref\ E2.SubEvent)$\\
  \hline\mbox{}&\\[-2ex]
  \multirow{5}{*}{possible coref.} 
                  & $(E1.Killer == E2.Killer)\ \&\ (E1.Time \sim E2.Time)$\\[1ex]
                  & $(E1.Killer == E2.Killer)\ \&\ (E1.Place \sim E2.Place)$\\[1ex]									
                  & $(E1.Killer == E2.Killer)\ \&\ (E1.Tool == E2.Tool)$\\[1ex]																		
                  & $(E1.Tool == E2.Tool)\ \&\ (E1.Time \sim E2.Time)$\\[1ex]
                  & $(E1.Place \sim E2.Place)\ \&\ (E1.Time \sim E2.Time)$\\
  \hline
  \end{tabular}
  
\subsection{Dying}

  \begin{tabular}{|l|c|}
  \hline
	\multirow{1}{*}{certain coref.} 
                  & $(E1.Protagonist == E2.Protagonist)$ \\[.5ex]
  \hline\mbox{}&\\[-2ex]
  \multirow{5}{*}{possible coref.} 
                  & $(E1.Place ~ E2.Place)\ \&\ (E1.Time \sim E2.Time)$\\[1ex]
                  & $(E1.Cause == E2.Cause)\ \&\ (E1.Time \sim E2.Time)$\\[1ex]									
                  & $(E1.Killer == E2.Killer)\ \&\ (E1.Tool == E2.Tool)$\\[1ex]																		
                  & $(E1.Tool == E2.Tool)\ \&\ (E1.Time \sim E2.Time)$\\[1ex]
                  & $(E1.Place \sim E2.Place)\ \&\ (E1.Time \sim E2.Time)$\\
  \hline
  \end{tabular}

\subsection{Charging}

  \begin{tabular}{|l|c|}
  \hline
	\multirow{1}{*}{certain coref.} 
                  & $(E1.Accused == E2.Accused)\ \&\ (E1.Time == E2.Time)$ \\[1ex]
  \hline\mbox{}&\\[-2ex]
  \multirow{3}{*}{possible coref.} 
                  & $(E1.Accused == E2.Accused)\ \&\ (E1.Charges == E2.Charges)$\\[1ex]
                  & $(E1.Arraign\_authority == E2.Arraign\_authority)\ \&$\\
                  & $(E1.Place == E2.Place)$\\[1ex]									
                  & $(E1.Place \sim E2.Place)\ \&\ (E1.Time \sim E2.Time)$\\
  \hline
  \end{tabular}

\subsection{Shooting}

  \begin{tabular}{|l|c|}
  \hline
	\multirow{2}{*}{certain coref.} 
                  & $(E1.Agent == E2.Agent)\ \&\ (E1.Goal == E2.Goal)\ \&$\\
                  & $(E1.Time == E2.Time)$ \\[1ex]
	                & $(E1.SubEvent\ hasCoref\ E2.SubEvent)$\\
  \hline\mbox{}&\\[-2ex]
  \multirow{4}{*}{possible coref.} 
                  & $(E1.Place \sim E2.Place)\ \&\ (E1.Time \sim E2.Time)$\\[1ex]
                  & $(E1.Agent == E2.Agent)\ \&\ (E1.Time == E2.Time) $\\[1ex]			                   & $(E1.Agent == E2.Agent)\ \&\ (E1.Goal == E2.Goal) $\\[1ex]
                  & $(E1.Projectile == E2.Projectile)\ \&\ (E1.Time == E2.Time)$\\
 \hline
  \end{tabular}

\subsection{Attacking}

  \begin{tabular}{|l|c|}
  \hline
	\multirow{2}{*}{certain coref.} 
                  & $(E1.Assailant == E2.Assailant)\ \&\ (E1.Victim == E2.Victim)\ \&$\\
                  & $\ (E1.Time == E2.Time)$ \\[1ex]
	              & $(E1.SubEvent\ hasCoref\ E2.SubEvent)$\\
  \hline\mbox{}&\\[-2ex]
  \multirow{4}{*}{possible coref.} 
                  & $(E1.Place \sim E2.Place)\ \&\ (E1.Time \sim E2.Time)$\\[1ex]
                  & $(E1.Assailant == E2.Assailant)\ \&\ (E1.Time == E2.Time) $\\[1ex]	                   & $(E1.Victim == E2.Victim)\ \&\ (E1.Time == E2.Time)$\\[1ex]
                  & $(E1.Weapon == E2.Weapon)\ \&\ (E1.Time == E2.Time)$\\
 \hline
  \end{tabular}
\newpage



\bibliographystyle{splncs}
\bibliography{bibliography}

\end{document}